\documentclass[sigconf]{acmart}

\usepackage{multirow}
\usepackage{graphicx}
\usepackage{subfigure}
\usepackage{makecell}

\usepackage{tablefootnote}

\usepackage{hyperref}
\makeatletter
\def\UrlAlphabet{%
      \do\a\do\b\do\c\do\d\do\e\do\f\do\g\do\h\do\i\do\j%
      \do\k\do\l\do\m\do\n\do\o\do\p\do\q\do\r\do\s\do\t%
      \do\u\do\v\do\w\do\x\do\y\do\z\do\A\do\B\do\C\do\D%
      \do\E\do\F\do\G\do\H\do\I\do\J\do\K\do\L\do\M\do\N%
      \do\O\do\P\do\Q\do\R\do\S\do\T\do\U\do\V\do\W\do\X%
      \do\Y\do\Z}
\def\UrlDigits{\do\1\do\2\do\3\do\4\do\5\do\6\do\7\do\8\do\9\do\0}
\g@addto@macro{\UrlBreaks}{\UrlOrds}
\g@addto@macro{\UrlBreaks}{\UrlAlphabet}
\g@addto@macro{\UrlBreaks}{\UrlDigits}

\usepackage{cleveref}
\crefname{section}{§}{§§}
\Crefname{section}{§}{§§}

\AtBeginDocument{%
  \providecommand\BibTeX{{%
    \normalfont B\kern-0.5em{\scshape i\kern-0.25em b}\kern-0.8em\TeX}}}
\renewcommand{\vec}[1]{\boldsymbol{#1}} 

\setcopyright{acmcopyright}

\copyrightyear{2022}
\acmYear{2022}
\setcopyright{acmcopyright}\acmConference[SIGIR '22]{Proceedings of the 45th
International ACM SIGIR Conference on Research and Development in Information
Retrieval}{July 11--15, 2022}{Madrid, Spain}
\acmBooktitle{Proceedings of the 45th International ACM SIGIR Conference on
Research and Development in Information Retrieval (SIGIR '22), July 11--15, 2022,
Madrid, Spain}
\acmPrice{15.00}
\acmDOI{10.1145/3477495.3536331}
\acmISBN{978-1-4503-8732-3/22/07}

\begin{document}

\title{A Low-Cost, Controllable and Interpretable Task-Oriented Chatbot: With Real-World After-Sale Services as Example}



\author{Xiangyu Xi$\dagger$}

\affiliation{%
  \institution{Meituan Group}
  \institution{National Engineering Research Center for Software Engineering, Peking University}
  \country{Beijing, China}
}
\email{xixy10@foxmail.com}

\author{Chenxu Lv$\dagger$}
\affiliation{%
  \institution{Meituan Group}
  \country{Beijing, China}
}
\email{lvchenxu@meituan.com}

\author{Yuncheng Hua}
\affiliation{%
  \institution{Meituan Group}
  \country{Beijing, China}
}
\email{huayuncheng@meituan.com}

\author{Wei Ye$\ast$}
\affiliation{%
  \institution{National Engineering Research Center for Software Engineering, Peking University}
  \country{Beijing, China}
}
\email{wye@pku.edu.cn}

\author{Chaobo Sun}
\affiliation{%
  \institution{Meituan Group}
  \country{Beijing, China}
}
\email{sunchaobo@meituan.com}

\author{Shuaipeng Liu}
\affiliation{%
  \institution{Meituan Group}
  \country{Beijing, China}
}
\email{liushuaipeng@meituan.com}

\author{Fan Yang}
\affiliation{%
  \institution{Meituan Group}
  \country{Beijing, China}
}
\email{yangfan79@meituan.com}

\author{Guanglu Wan}
\affiliation{%
  \institution{Meituan Group}
  \country{Beijing, China}
}
\email{wanguanglu@meituan.com}





\thanks{$\dagger$ The first two authors contributed equally.}
\thanks{$\ast$ Wei Ye is the corresponding author.}

\renewcommand{\shortauthors}{Xiangyu, Chenxu and Yuncheng, et al.}


%
\begin{abstract}

Though widely used in industry, traditional task-oriented dialogue systems suffer from three bottlenecks: (i) difficult ontology construction (e.g., intents and slots); (ii) poor controllability and interpretability; (iii) annotation-hungry. In this paper, we propose to represent utterance with a simpler concept named Dialogue Action, upon which we construct a tree-structured TaskFlow and further build task-oriented chatbot with TaskFlow as core component. A framework is presented to automatically construct TaskFlow from large-scale dialogues and deploy online. Our experiments on real-world after-sale customer services show TaskFlow can satisfy the major needs, as well as reduce the developer burden effectively.

\end{abstract}




\maketitle

\section{Introduction}

Task-oriented chatbot, which aims to assist users in completing certain tasks, has been proven valuable for real-word business especially after-sale customer services.
\cite{li2017alime,zhu2019case,acharya-etal-2021-alexa,sun-etal-2021-adding}
A well-designed chatbot can help standardize the service process, as well as alleviate the pressure of after-sale staffs.






Traditional task-oriented dialogue systems require the domain experts to manually develop a structured ontology (e.g., intents and slots) as foundations, and then build modules including Natural Language Understanding (NLU), Dialog State Tracking (DST), Dialog Policy (DP), and Natural Language Generation (NLG) respectively.
However, in industrial practice, such system design and construction method suffer from three bottlenecks. 
(i) It's difficult to represent real-world complex utterances with combination of dialogue acts and slots. For example, our preliminary exploration shows that it takes more than 20 slots and 100 possible values to fully express user's semantic information for a after-sale customer service.
(ii) Most dialogue policy works exploit an implicit manner, which leads to unsatisfactory controllability and interpretability for industrial systems.
(iii) Existing works adopt supervised learning paradigm which heavily relies on human-annotated data, while the labeling process can be costly and error-prone.

To address above challenges, inspired by the speech act theory \cite{searle1975taxonomy,traum1992conversation}, we propose to represent utterance with a simpler concept, named Dialogue Action, instead of dialogue acts and slots.
Dialogue action is regarded as utterances with unique and identical semantic information within the dialogues corpus, and can be automatically obtained by clustering.
In this way, most utterances can be represented by a specific dialogue action.
For controllability and interpretability, we exploit an explicit manner and build task-oriented chatbot based on TaskFlow, which is a tree structure with dialogue actions as nodes and dialogue action transition as edges.
Finally, we present a framework to automatically construct TaskFlow from large-scale dialogues and deploy online, to further reduce the developer burden.
\begin{figure*}[!htb]
    \centering
    \includegraphics[scale=0.25]{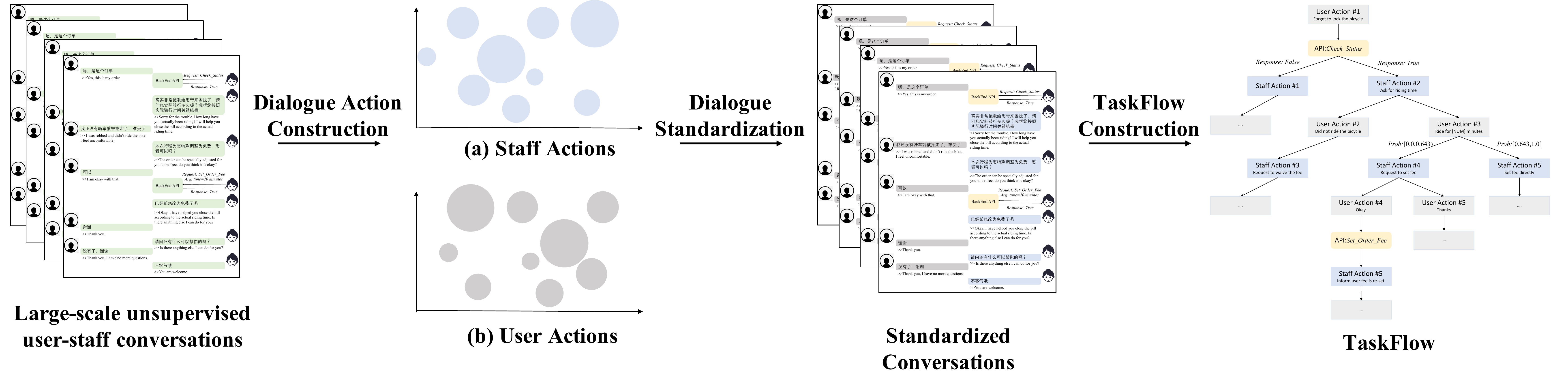}
    \caption{The offline part. 
    Staff- and user-related information is marked by light blue and grey respectively. Note that the figures are for illustration only and more details can be found in the following sections.}
    \label{fig:model}
\end{figure*}
The contributions of our work are:
\begin{enumerate}
    \item We propose to build task-oriented chatbot based on dialogue actions and TaskFlow, which has the advantage of simplicity, controllability and interpretability.
    \item We present an effective framework to automatically construct TaskFlow from large-scale dialogues and deploy the TaskFlow online.
    \item Our exploration on real-world after-sale customer services shows that the TaskFlow can satisfy the majority of user needs, as well as effectively reduce the developer burden.
\end{enumerate}

\section{Overview}


Our proposed framework mainly consists of two parts, named offline part and online part. We exemplify using a Chinese after-sale customer service of electric bike rental business, where users and staffs communicate online through text messages.
With the offline part, we automatically construct TaskFlow from large scale chat logs.
After that, with the online part, the TaskFlow can be rapidly deployed online and work as the core component of task-oriented chatbot. The details of both parts are introduced in the following.


\subsection{Offline Part}
As Figure \ref{fig:model} shows, 3 carefully-designed steps are performed sequentially in the offline part: 1) Dialogue action construction, which constructs dialogue actions for user/staff by clustering; 2) Dialogue standardization, which standardizes the dialogues by mapping each utterance to a dialogue action; 3) TaskFlow construction, which constructs TaskFlow with the standardized dialogues.

\subsubsection{Dialogue Action Construction}
\label{sec:action_construction}
A dialogue action is regarded as a group of utterances with unique and identical semantic information. Inspired by \citet{lv-etal-2021-task-oriented}, we exploit a popular two-stage method to cluster utterances from large scale dialogues as follows.

\textbf{Feature Extractor} Sentence representation generated by pre-trained language models has been widely used as features for clustering. In this paper, we exploit ConSERT \cite{yan-etal-2021-consert}, which solves the collapse issue of BERT-derived sentence representations by contrastive learning, as feature extractor. The ConSERT\footnote{https://github.com/yym6472/ConSERT} is fine-tuned with the dialogues data, to make the sentence representation more task-oriented and applicable to clustering.  The output of {\tt [CLS]} token is utilized as feature of each utterance for further clustering.

\textbf{Clustering} We use K-means \cite{krishna1999genetic} to group utterances, and each cluster is treated as a dialogue action. For purity, annotator may check and slightly modify the clusters. Our practice shows the clusters are of high quality and require limited human efforts. 

\subsubsection{Dialogue Standardization}
\label{sec:Standardization}


%
Dialogue standardization aims to standardize the dialogue by mapping each utterance to a specific dialogue action. 
Inspired by \citet{yu-etal-2021-shot}, we exploit a retrieval-based method, which retrieves clustered utterances that are most similar to the given input utterance, and label the input based on corresponding clusters. Compared with traditional classification methods \cite{kowsari2019text}, such design has two advantages: 1) retrieval-based method is more suitable for our scenario where instance numbers of each cluster are imbalanced. 2) retrieval-based method can adapt to modification of actions (e.g. create new action or remove existing action) without having to retrain the model.

\begin{figure}[htb]
    \centering
    \includegraphics[scale=0.32]{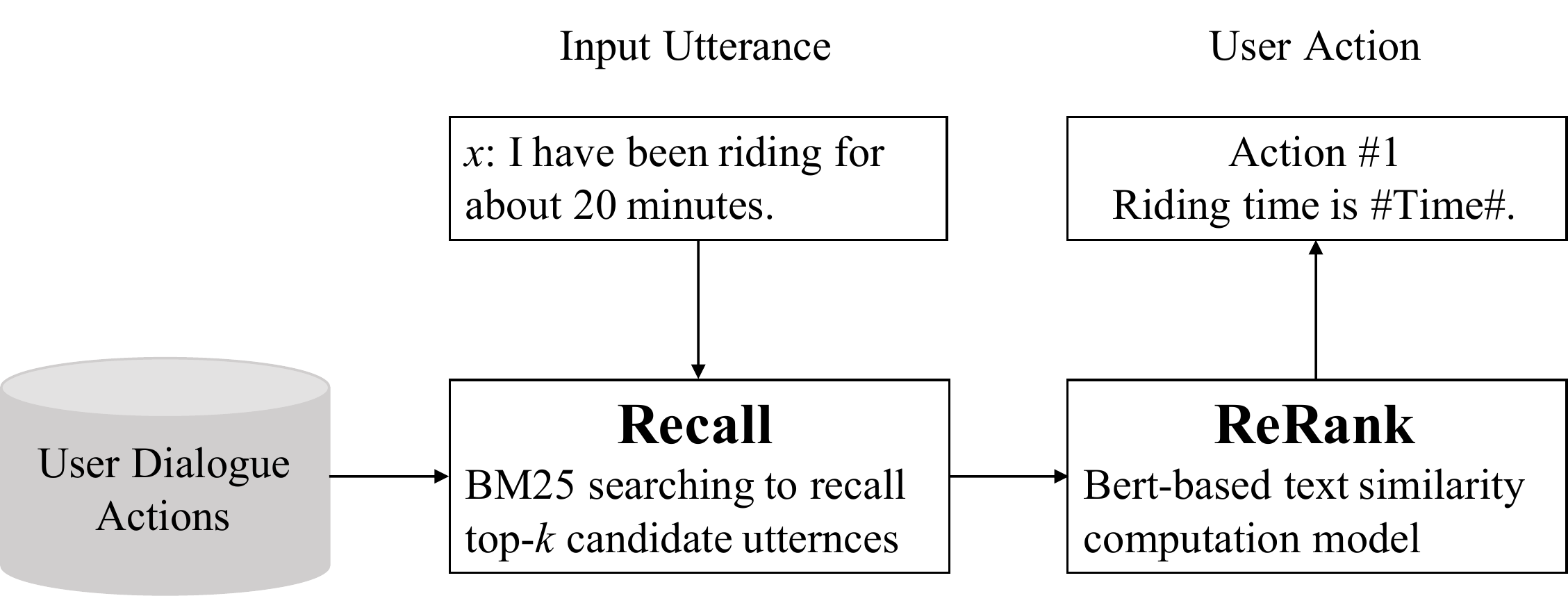}
    \caption{The workflow of retrieval-based method.}
    \label{fig:retrieval}
\end{figure}

Specifically, as Figure \ref{fig:retrieval} shows, given an input utterance $x$, we first use BM25 algorithm \cite{robertson2009probabilistic} to recall top $k$ utterances $\{x_1,...,x_k\}$ from all clustered utterances.
Further, we exploit a BERT-based text similarity computation model $S$ to rerank the utterances and select the utterance $\hat x$ with highest similarity to $x$, which can be denoted by:
\begin{equation}
    \hat x = \underset{x_i \in \{x_1,...,x_k\}}{\mathrm{argmax}} S(x,x_i)
\end{equation}
where $S(x,x_i)$ denotes the similarity between $x$ and $x_i$. 
The text similarity computation model is based on BERT, and takes concatenation of $x$ and $x_i$ as inputs (The input sentence is ({\tt [CLS]}, $x$, {\tt [SEP]}, $x_i$)). It first encodes the concatenated utterances into continuous representations, and computes similarity with output of the {\tt [CLS]} token (denoted by $\vec{h}_1$ in the following) as follows:

\begin{equation}
\begin{aligned}
    (\vec{h}_1,...,\vec{h}_n) &= {\rm BERT}([x; x_i]) \\
    S(x,x_i) &= \rm Sigmoid (\vec{W} \vec{h}_1 + b)
\end{aligned}
\end{equation}
where $\vec{W}$ and $b$ are trainable weight parameters.

\subsubsection{TaskFlow Construction}
\begin{figure}
    \centering
    \includegraphics[scale=0.25]{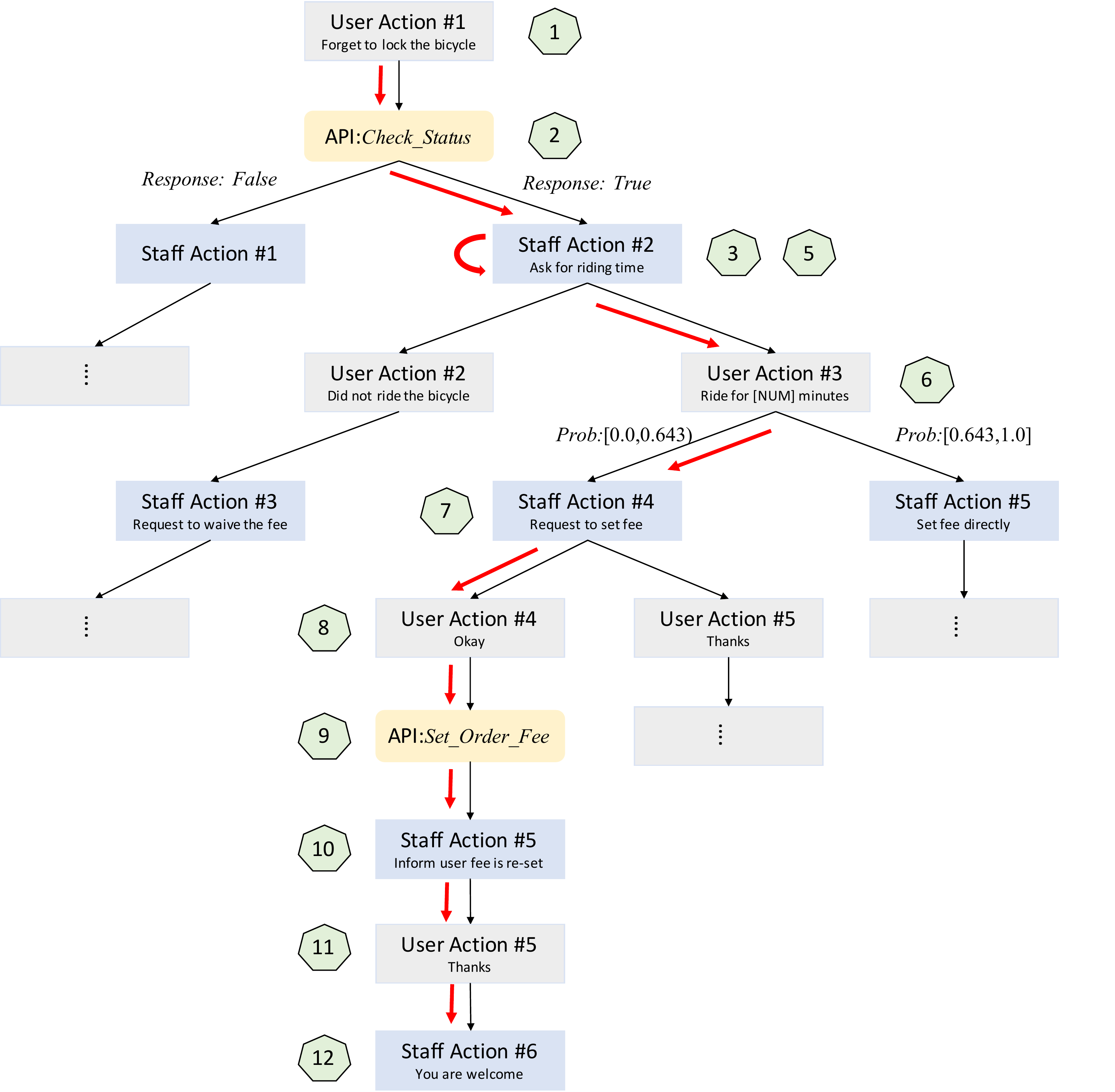}
    \caption{A partial TaskFlow}
    \label{fig:taskflow}
\end{figure}

TaskFlow is a tree with dialogue action as nodes, and the edges between nodes describe how the conversation proceeds (Refer to Figure \ref{fig:taskflow} for a TaskFlow sample). We construct TaskFlow from the standardized dialogues, each of which can be treated as a sequence of user/staff actions.
It's intuitive to construct TaskFlow by directly inserting action sequence of each dialogue into a tree. 
However, due to the dynamic nature of dialogues, the action distribution across the corpus would be too scattered, given a certain turn. 
Similar conversation fragments may lie in different turns of dialogues.
Thus, instead of the direct way, we first build a N-gram model of dialogue actions which captures the local conversation pattern more accurately.
Further, we sample high-quality action sequences from N-gram model, and the sampled action sequences are merged together, forming the TaskFlow.



\textbf{Building N-gram Model} Given an action sequence $A = \{a_1,a_2,...,a_n\}$, the probability can be approximated as follows:
\begin{equation}
\begin{aligned}
    P(A) &= \prod_{i=1}^{n}P(a_i|a_{1:i-1}) \\
    &\approx \prod_{i=1}^{n}P(a_i|a_{i-N-1:i-1})
\end{aligned}
\end{equation}
where $a_{i:j}=\{a_i,a_{i+1},...,a_j\}$ denotes the subsequence with $i$ as starting index and $j$ as ending index.
Following maximum likelihod estimation, we compute the N-gram as:
\begin{equation}
    P(a_i|a_{i-N-1:i-1}) = \frac{C(a_{i-N-1:i})}{C(a_{i-N-1:i-1})}
\end{equation}
where $C(a_{i:j})$ denotes the count of $a_{i:j}$ in corpus.

\textbf{Sampling Action Sequence} We sample action sequences based on the N-gram model. Specifically, the sampling process starts with {\tt [SOS]} and ends with {\tt [EOS]}. 
We follow the beam search method \cite{freitag2017beam} but with a different strategy. In each step, we extend every partial action sequence in the beam with its top K actions. Once the {\tt [EOS]} symbol is appended to a partial action sequence, it is removed from the beam, and a complete action sequence is sampled.

\textbf{Generating TaskFlow} TaskFlow is generated by simply inserting the sampled action sequences into a tree individually (sequence by sequence). The transition probability is recorded as the condition on the edge between action nodes.

\textbf{Post-Processing} 
With the generated TaskFlow, operational staff can easily determine 1) which action nodes require API calls (e.g., staffs need to lock the bike remotely), 2) which conditions on the edges require modification (e.g., API response be specific value).
For example, if a specific user action node has multiple children (i.e., staff action nodes) with different semantic information, an API call is probably required and each edge may correspond to different API response. 
We manually add API call nodes and modify the conditions on the edge, if necessary. In this way, TaskFlow consists of three types of nodes, named user action node, staff action node and api call node respectively.

\subsection{Online Part}
\label{sec:online}

To deploy the TaskFlow online and enable interaction with users, we build an execution engine, based on the principle that an ideal conversation should follow the paths in TaskFlow. The execution engine stores and updates the path corresponding to the current conversation.
Basically, it moves along the path if the condition on an edge is satisfied, and respond to users when encountering staff action node. 
Given a new user utterance, the execution engine first categorizes the utterance into an user action with the retrieval-based model from Section \ref{sec:Standardization}.
Then, if the user action node's children are staff action nodes, a staff action node is selected based on the conditions on the edges. 
Otherwise, the user action node has only one API call node as child. The engine first uses a Parameter Value Extraction Module (introduced in the following) to extract required parameter values, and then executes the API call node. 
Further, the execution engine moves along the path until it reaches a leaf node or the conversation is closed.


\begin{figure}[htb]
    \centering
    \includegraphics[scale=0.33]{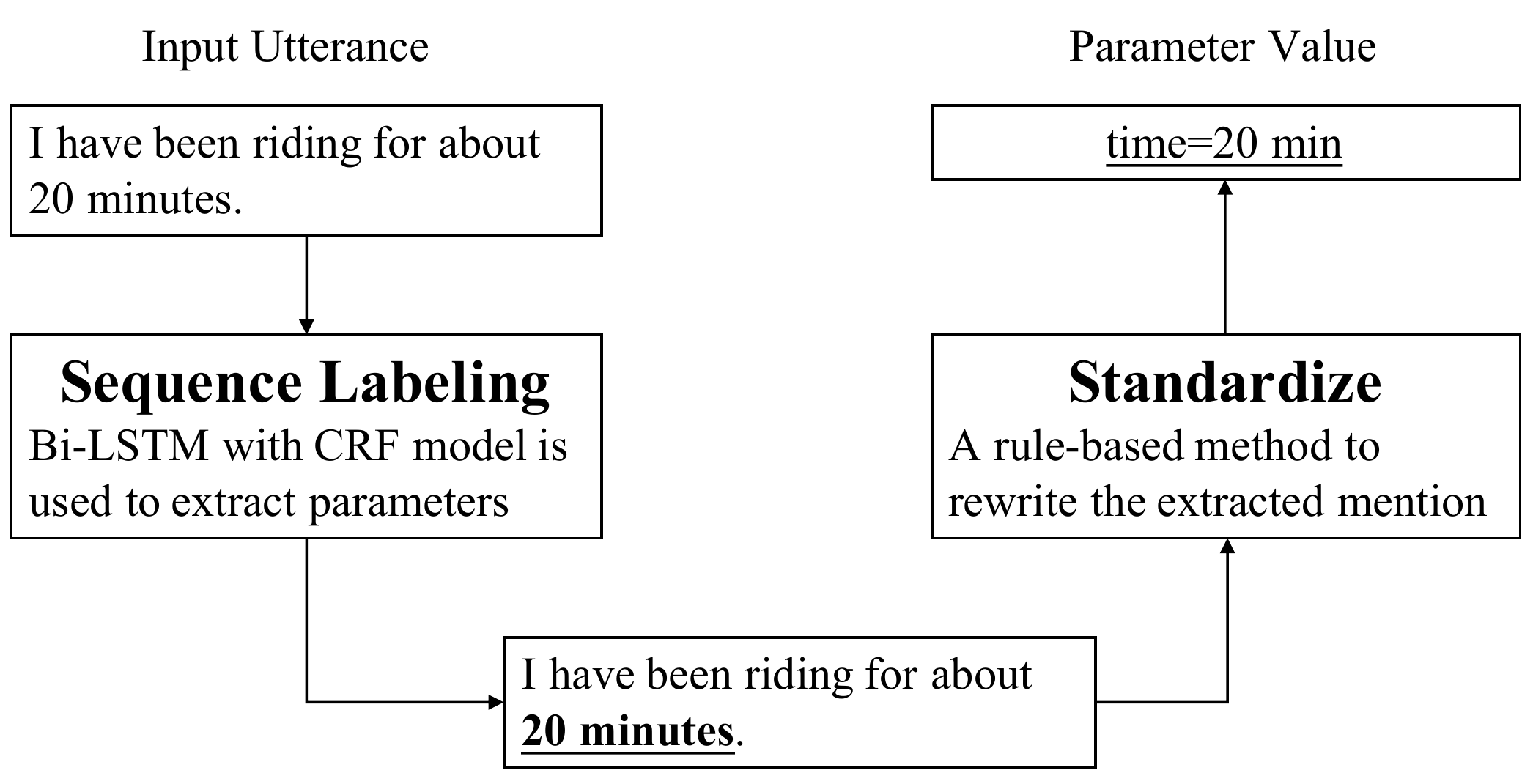}
    \caption{Workflow of Parameter Value Extraction Module.}
    \label{fig:parameter}
\end{figure}

\textbf{Parameter Value Extraction Module} The parameter value extraction module extracts parameter values for the API call from user utterances (e.g., time, user id). Specifically, we use a two-stage method to extract parameters as Figure \ref{fig:parameter} shows. The Bi-LSTM with CRF model \cite{lample2016neural,xiangyu2019hybrid,xi2021improving} is first used to perform sequence labeling and extract the mention containing the parameter value. Then a rule-based method is utilized to rewrite the mention, and get the standardized parameter value.

\section{INDUSTRIAL APPLICATION}

\subsection{Online Deployment}
TaskFlow is integrated into our online chatbot in the following three typical scenarios of after-sale customer service of electric bike rental:
\begin{enumerate}
    \item \emph{Forget\_to\_Lock\_bike} where customers finished riding but forgot to lock the bike. The customers may require the staff to remotely lock the bike and reduce the fees.
    \item \emph{Mechanical\_Failure} where customers encountered mechanical failures such as brake failure during the ride. The customers may claim for refund, due to poor user experience.
    \item \emph{Out\_Of\_Power} where the electric bike ran out of power during the ride. The customers may also claim for refund.
\end{enumerate}
All the scenarios involve strict and complex business logic. For example, the staffs need to judge whether the fee can be waived by checking back-end APIs.
We construct TaskFlow for each scenario individually.
Specifically, we sample 50,000 utterances from the corpus, and grouped them into 100 clusters respectively. After manual modification, 82 user actions and 93 staff actions are retained.
The text similarity computation model adopt BERT$_{\rm BASE}$ model as encoder.
We build a 4-gram model and extend action sequence in the beam with its top 5 actions. 

\subsection{Online Evaluation}

\subsubsection{Evaluation Metrics}
We randomly sample 150 dialogues for each scenario. Annotators with domain knowledge are asked to grade each dialogue by ``-1'', ``0'' or ``1'', and the grading criteria can be summarized as follows:
(i) Score ``-1'' denotes that Chatbot can not handle user requirements correctly.
(ii) Score ``0'' denotes that Chatbot can handle user requirements correctly, but may generate influent or incomplete response.
(iii) Score ``1'' denotes that Chatbot can handle user requirements correctly and perfectly complete the conversation.
\begin{table}[hbt]
    \centering
    \caption{Statistical Results of Human Evaluation (\%)}
    \begin{tabular}{lccc}
    \hline
    Scenario &  -1 & 0 & 1\\
    \hline
    \emph{Forget\_to\_Lock\_bike} & 7.69  & 12.82 & 79.49 \\
    \emph{Mechanical\_Failure} & 14.20   & 10.56 & 75.24 \\
    \emph{Out\_Of\_Power} & 9.34   & 9.41 & 81.25 \\
    \hline
    \end{tabular}
    
    \label{tab:human_evaluation}
\end{table}


\subsubsection{Human Evaluation}
The statistical results of human evaluation is shown in Table \ref{tab:human_evaluation}, from which we can observe that:
(1) TaskFlow has satisfactory performance in terms of understanding and meeting user requirements, which is the core function of task-oriented chatbot.
For exmaple, in scenario \emph{Forget\_to\_Lock\_bike}, TaskFlow can correctly handle user requirements of 96.31\% dialogues. 
(2) Due to diversified user expressions in real-world scenarios, TaskFlow may generate incomplete and influent responses (e.g., 12.82\% of dialogues in \emph{Forget\_to\_Lock\_bike}). For example, users may complain or curse about unexpected brake failures.

\subsection{In-Depth Analysis}

\subsubsection{Analysis of Efficiency Issue}
To quantitatively explore the low-cost characteristic of our method, we build a traditional task-oriented chatbot in \emph{Forget\_to\_Lock\_bike} (denoted by Traditional in the following) following \citet{yao2013recurrent}, and compare the required human efforts in Table \ref{tab:comparison}. Though achieving comparable performance, our TaskFlow is low-cost and require much less human efforts (i.e., 9 v.s. 21 person-days).

Specifically, the building process can be divided into four steps, and person-days (abbr. as pds) required by each step is recorded in Table \ref{tab:comparison}.
Since traditional system involves multiple modules, it requires large-scale data annotation and more person-days for training and deploying online.
With the presented automatic framework, TaskFlow-based system requires much less annotation and human efforts, thus effectively reducing developer burden.

\begin{table}[hbt]
    \centering
    \caption{Human efforts required for different chatbots.}
    \begin{tabular}{llrr}
    \hline
    \textbf{Step} & \textbf{Process} & Traditional & TaskFlow\\
    \hline 
    1 & Ontology Construction & 3 pds & 1 pds\\
    2 & Data Annotation & 12 pds & 4 pds\\
    3 & Training Model & 3 pds& 2 pds\\
    4 & Online Deployment & 3 pds& 2 pds\\
    \hline
    & Total & 21 pds & 9 pds \\
    \hline
    \end{tabular}
    \label{tab:comparison}
\end{table}

\subsubsection{Analysis of Controllability and Interpretability}

\begin{figure}
    \centering
    \includegraphics[scale=0.25]{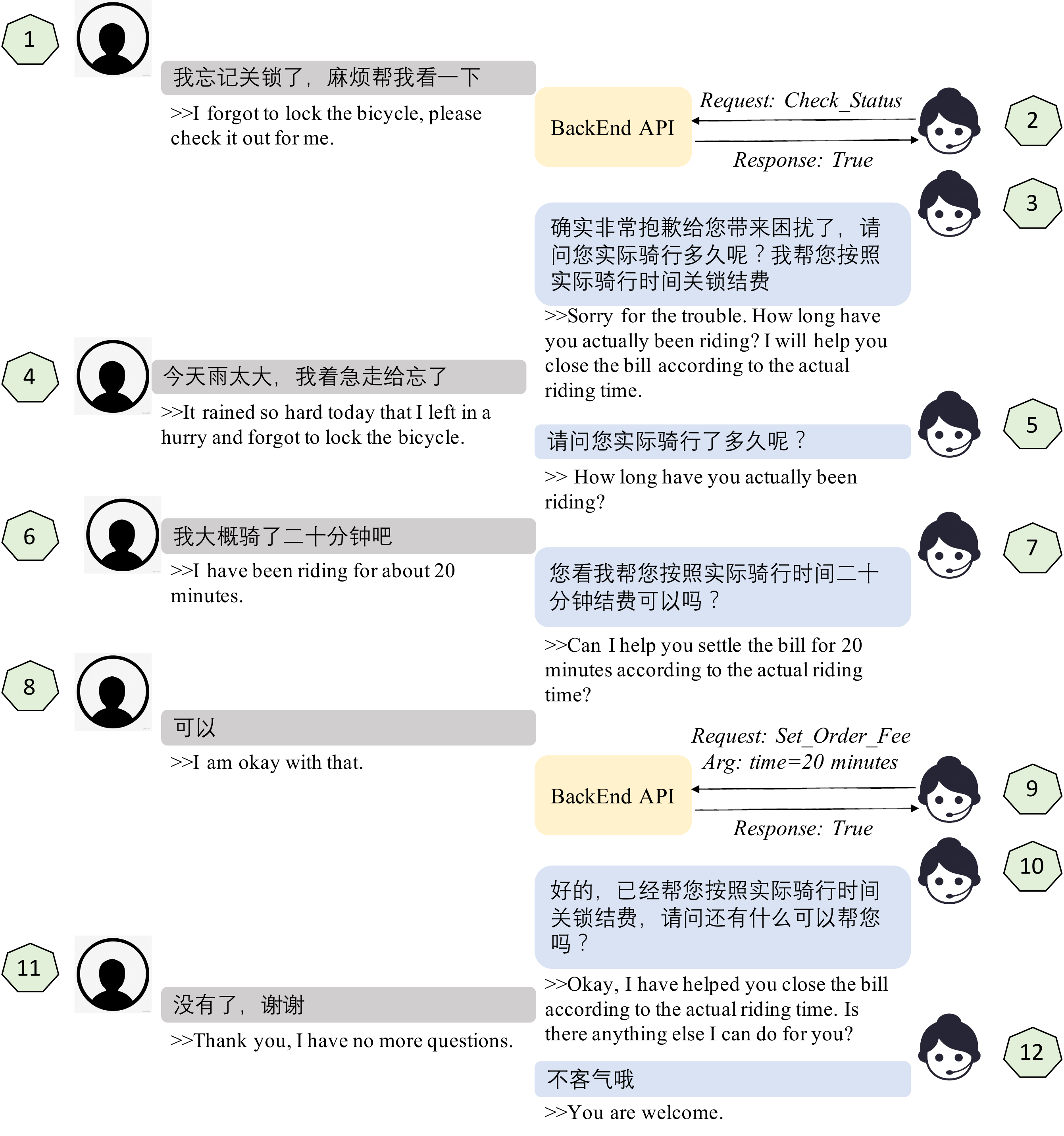}
    \caption{Dialogue by interacting with TaskFlow in Figure \ref{fig:taskflow}}
    \label{fig:case}
\end{figure}

With an explicit manner, TaskFlow-based system naturally has the advantage of controllability and Interpretability. 
Specifically, conversations exactly follow the paths in TaskFlow while traditional dialogue policy module may generate unexpected dialogue act. 
We present the TaskFlow in \emph{Forget\_to\_Lock\_bike} and a concrete conversation in Figure \ref{fig:case}. 
If the condition on edge is met, the execution engine will move along the edge and respond according to the staff action node. For example, with API {\tt Check\_Status} returning {\tt True}, the engine moves to staff action \#2 node and respond with 3rd utterance.
In this way, unexpected response will never be generated, and the path clearly shows how the conversation is completed.

The above advantages also lead to better flexibility. 
TaskFlow can be easily modified while it's hard to manually intervene traditional dialogue system.
When encountering changes of customer service policy, traditional dialogue systems may be unavailable, and we need to retrain the model with new labeled data.
In contrast, our TaskFlow can be edited and adapt to changes immediately (without training). 

\section{Conclusion}

In this paper, to meet the simplicity, controllability and interpretability required by industrial dialogue systems, we propose a framework to build task-oriented chatbots based on dialogue actions and TaskFlow, from large-scale dialogues. The experiments show such a framework can effectively satisfy majority needs and reduce human efforts.

In the future, we are interested in optimizing the framework overally especially exploring more sophisticated TaskFlow construction methods. Besides, since we construct TaskFlow for each scenario separately, how to construct multi-scenario TaskFlow is worth investigating. 

\section{RELEVANCY TO SIRIP 2022 THEMES}

In this paper, we propose a framework to build task-oriented chatbots based on dialogue actions and TaskFlow, to meet the simplicity, controllability and interpretability required by industrial systems.
Our practical experience shows that such a framework can effectively satisfy majority needs and reduce human efforts.
The retrieval system greatly contributes to the success of TaskFlow.

\section{Presenter BIOGRAPHY}
Presenter: Xiangyu Xi. He is an algorithm engineer at Meituan. His research interests include information retrieval, dialogue system, and information extraction.
He is currently working on the construction of task-oriented dialogue system in Meituan.

\section{Company Portrait}
Meituan is China's leading shopping platform for locally found consumer products and retail services including entertainment, dining, delivery, travel and other services.

\bibliographystyle{ACM-Reference-Format}
\bibliography{sample-base}

\appendix

\end{document}